\documentclass[11pt]{article}

\usepackage{natbib}
\usepackage{graphicx}
\usepackage{booktabs}
\usepackage{tabularx}
\usepackage[margin=1.1in]{geometry}
\usepackage{microtype}
\usepackage{xcolor}
\usepackage{enumitem}
\usepackage{eso-pic}
\usepackage{needspace}
\usepackage[colorlinks=true,citecolor=blue,urlcolor=blue,linkcolor=blue]{hyperref}
\microtypesetup{expansion=false}
\usepackage{listings}
\BeforeBeginEnvironment{lstlisting}{\Needspace{12\baselineskip}}
\definecolor{pykeyword}{RGB}{0,92,175}
\definecolor{pystring}{RGB}{163,21,21}
\definecolor{pycomment}{RGB}{0,128,0}
\definecolor{pybg}{RGB}{248,248,248}
\title{jNO: A JAX Library for Neural Operator and Foundation Model Training}
\author{Leon Armbruster\thanks{Corresponding author. Email: \href{mailto:leon.armbruster@iisb.fraunhofer.de}{leon.armbruster@iisb.fraunhofer.de}} \and
  Rathan Ramesh \and
  Georg Kruse \and
  Christopher Straub\\[0.5em]
  Fraunhofer Institute for Integrated Systems and Device Technology IISB, Germany
}
\date{\today}

\AddToShipoutPictureFG*{%
  \AtPageUpperLeft{%
    \hspace*{\dimexpr\paperwidth-4.8cm\relax}%
    \raisebox{-2.2cm}[0pt][0pt]{\includegraphics[width=0.18\textwidth]{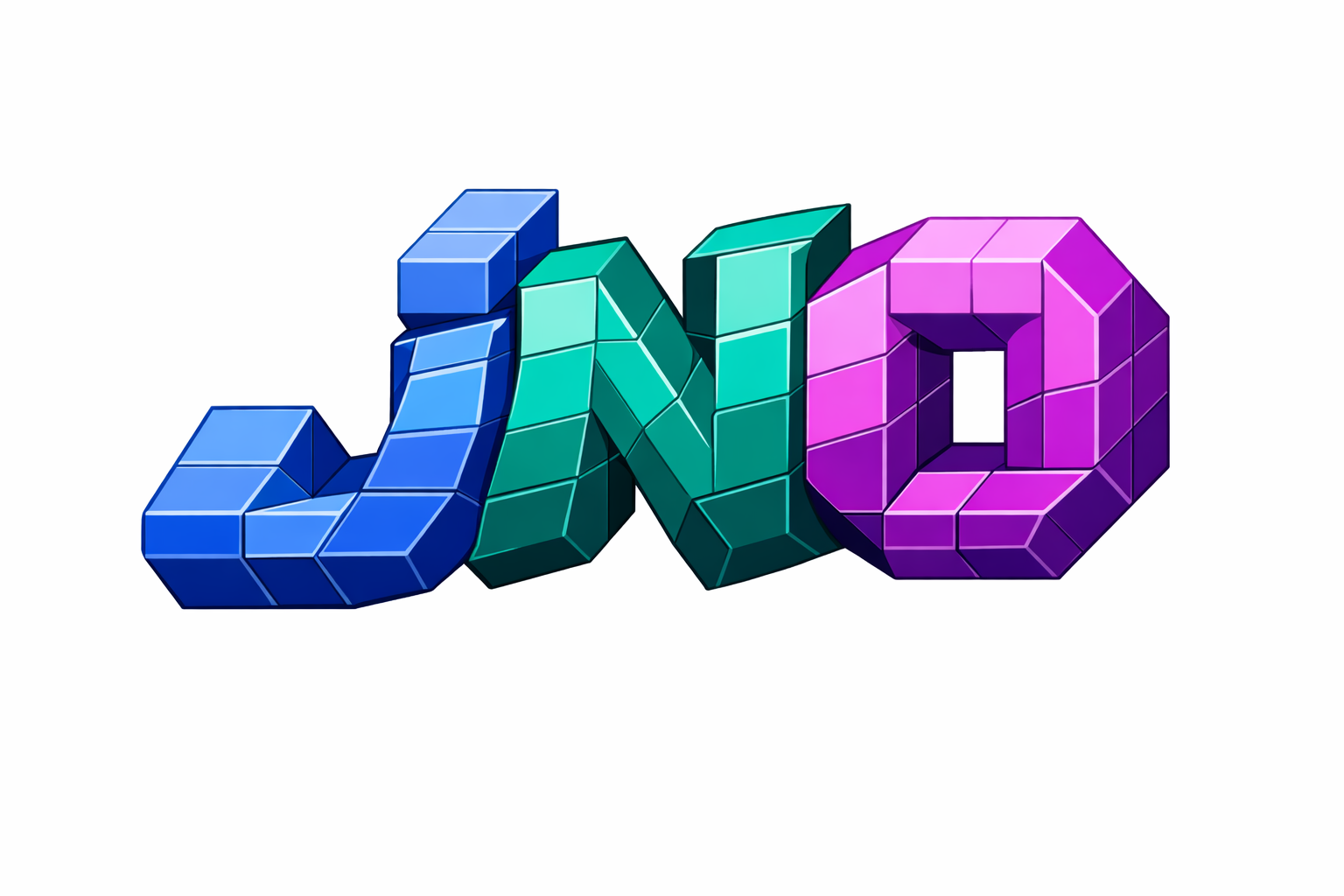}}%
  }%
}

\begin{document}

\maketitle

\begin{abstract}

  jNO (jax Neural Operators) is a JAX-native library for neural operators and foundation models with unified support for both data-driven and physics-informed training. Its core design is a tracing system in which domains, model calls, residuals, supervised losses, and diagnostics are written in one symbolic language and compiled into one optimization pipeline. This allows users to move between operator regression, mesh-aware residual evaluation, and PDE-constrained training without restructuring the surrounding code. 
  jNO also supports multi-model compositions, fine-grained control at parameter level (model, optimizer, and learning rate), hyperparameter tuning, and JAX-native workflows for translated PDE foundation-model families. The source repository is available at \url{https://github.com/FhG-IISB/jNO}.

\end{abstract}


\section{Introduction}

Neural operators are an extension of neural networks for learning mappings between infinite-dimensional function spaces, for instance, between initial and final states of complex physical systems~\citep{azizzadenesheli2024neural}. They are increasingly used in scientific machine learning for surrogate modeling, inverse problems, and PDE-constrained learning \citep{li2021fno,lu2021deeponet}. In practice, however, the surrounding software workflow is often fragmented across separate abstractions for geometry, loss construction, model definition, and training logic. This fragmentation becomes more pronounced when users want to combine supervised operator learning with physics-informed residuals~\citep{raissi2019physics} or transfer a pretrained backbone into a new PDE setting. jNO addresses this by treating the entire workflow as one traced program: users define domains and variables, compose model and differential expressions, and solve everything through a single interface.

This approach is inspired by compact equation-centric software such as DeepXDE \citep{lu2021deepxde}, while shifting the focus toward neural operator workflows \citep{li2021fno,lu2021deeponet} in a JAX-native stack \citep{bradbury2018jax}. A practical motivation is ecosystem balance: much foundation-model and operator tooling has been concentrated in PyTorch-first libraries. jNO is an attempt to bring this side of research into JAX with comparable usability.

jNO therefore targets two connected use cases: (1) unified neural operator and foundation-model training that supports data-driven supervision and physics-informed residuals in the same workflow; and (2) transfer or fine-tuning workflows in JAX with fine-grained parameter-level control over the neural networks.

\subsection{Foundation models and ecosystem consolidation}

\begin{figure}[htbp]
  \centering
  \includegraphics[width=1\textwidth]{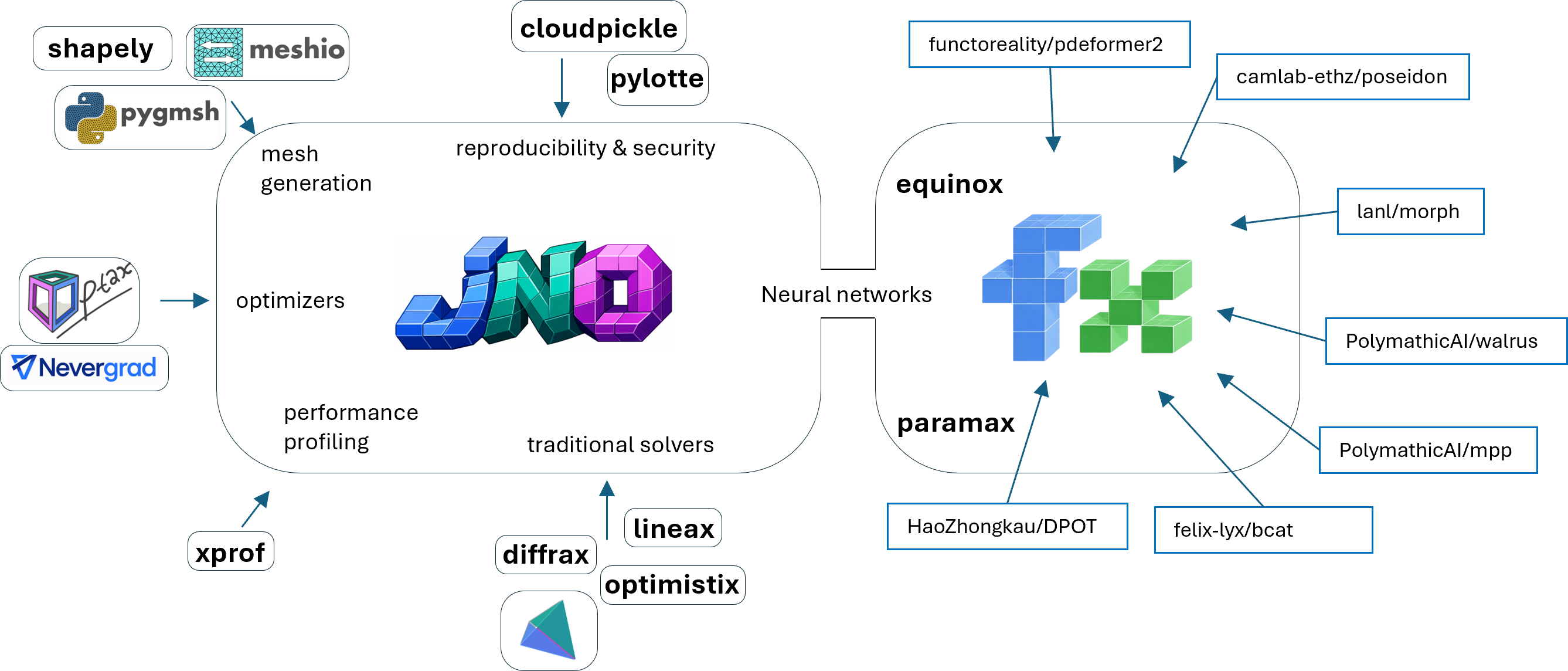}
  \caption{Overview of jNO's interdependencies.}
  \label{fig:jno-overview}
\end{figure}

A critical limitation in current foundation-model research is fragmentation: state-of-the-art PDE foundation models such as Poseidon \citep{herde2024poseidon}, Walrus \citep{mccabe2025walrus}, PDEformer2 \citep{pdeformer2}, MPP \citep{mccabe2024multiple}, Morph \citep{rautela2025morph}, BCAT \citep{liu2025bcat} and DPOT \citep{hao2024dpot} are maintained in separate repositories, predominantly in PyTorch ecosystems. This separation impedes systematic comparison, transfer learning workflows, and reproducible fine-tuning across model families. jNO acts as a unified JAX-native backend that consolidates these models into one framework, enabling researchers to evaluate and compose foundation models side-by-side without maintaining separate codebases. This consolidation is not merely organizational: it allows fine-grained parameter control, shared optimization interfaces, and hybrid objectives (e.g., combining a pretrained backbone with custom physics constraints) all within a single traced execution pipeline. Figure~\ref{fig:jno-overview} provides a high-level overview of jNO's ecosystem and interdependencies, while

\subsection{JAX tracing paradigm and performance benefits}

jNO's traced-symbolic design naturally complements JAX's own lazy tracing and XLA compilation strategy. Rather than introducing a new execution model, jNO extends JAX's functional programming paradigm by deferring both model calls and differential operators until compilation time. This alignment provides two concrete advantages. First, traced equations compile into a single XLA program where common sub-expressions are eliminated uniformly across objectives, operator definitions, and physics constraints~\citep{bradbury2018jax}. Second, the same traced abstraction supports efficient multi-device execution, gradient checkpointing, and fused optimization steps without requiring users to restructure their code. Compared to PyTorch-centric alternatives, this functional design and native JIT integration reduce compilation overhead and improve numerical stability through explicit XLA-level optimizations, while maintaining the prototyping simplicity critical for research iteration.

\subsection{Main features}

The main features highlighted in this publication are:

\begin{itemize}
  \item \textbf{Tracing-first design}: domains, models, residuals, and data terms are written in one symbolic language.
  \item \textbf{Mesh pipeline via PyGmsh/Gmsh}: mesh generation and loading are integrated into the domain API \citep{schlomer2018pygmsh,geuzaine2009gmsh}, supporting mesh-centric workflows aligned with FDM/FEM/FVM-style practice.
  \item \textbf{Fine-grained training controls}: model, optimizer, and learning-rate behavior can be configured at parameter level, alongside selective parameter masking and LoRA-style adaptation, without rewriting the equation-level program.
  \item \textbf{Foundation models in JAX}: jNO includes support for translated PDE foundation model families such as Poseidon, Walrus and Morph (maintained in foundax~\citep{foundax}).
  \item \textbf{FEM integration}: jNO extends its traced symbolic interface to variational formulations, enabling tagged weak form assembly for linear, nonlinear finite element adn transient problem workflows within the same domain and execution pipeline, while also supporting variational PINN type formulations through the same interface~\citep{kharazmi2019variationalphysicsinformedneuralnetworks}.
\end{itemize}

\subsection{Related work}

Several libraries address PDE learning in Python, each with different emphases. 
A high-level comparison is given in Table~\ref{tab:related_work}. 

\begin{table}[htbp]
  \centering
  \caption{Comparison of jNO with related PDE learning software}
  \label{tab:related_work}
  \small
  \begin{tabular}{p{3cm} p{3cm} p{3cm} p{3cm}}
    \toprule
    \textbf{Library}   & \textbf{Backend} & \textbf{Primary focus}            & \textbf{Operator emphasis}           \\
    \midrule
    DeepXDE            & TF, PyTorch, JAX & PINN                      & Moderate                              \\
    JAX-PI             & JAX              & PINN                 & Minimal                               \\
    NVIDIA PhysicsNeMo & PyTorch          & PINN + operator                   & Strong                                \\
    NeuralPDE.jl       & Julia            & PINN + operator                     & Moderate                                \\
    \textbf{jNO}       & JAX              & Neural operators and hybrid physics & Strong \\
    \bottomrule
  \end{tabular}
\end{table}

DeepXDE \citep{lu2021deepxde} is the most commonly used framework for compact PINN (Physics-Informed Neural Network~\citep{raissi2019physics}) software design. JAX-PI \citep{predictive2024jaxpi} is a JAX-native alternative focused specifically on PINN workflows. NVIDIA PhysicsNeMo \citep{hennigh2021nvidia} and NeuralPDE.jl \citep{zubov2021neuralpde} provide complementary strengths. jNO differs by centering on a single traced language for neural operators and hybrid objectives in JAX. Rather than separating operator learning, physics constraints, and mesh-aware data handling into distinct APIs, it exposes them through the same symbolic layer. It also places stronger emphasis on JAX-native translation and adaptation of PDE foundation model families~\citep{foundax}.

\section{Implementation}\label{sc:implementation}

\begin{figure}[htbp]
  \centering
  \includegraphics[width=1\textwidth]{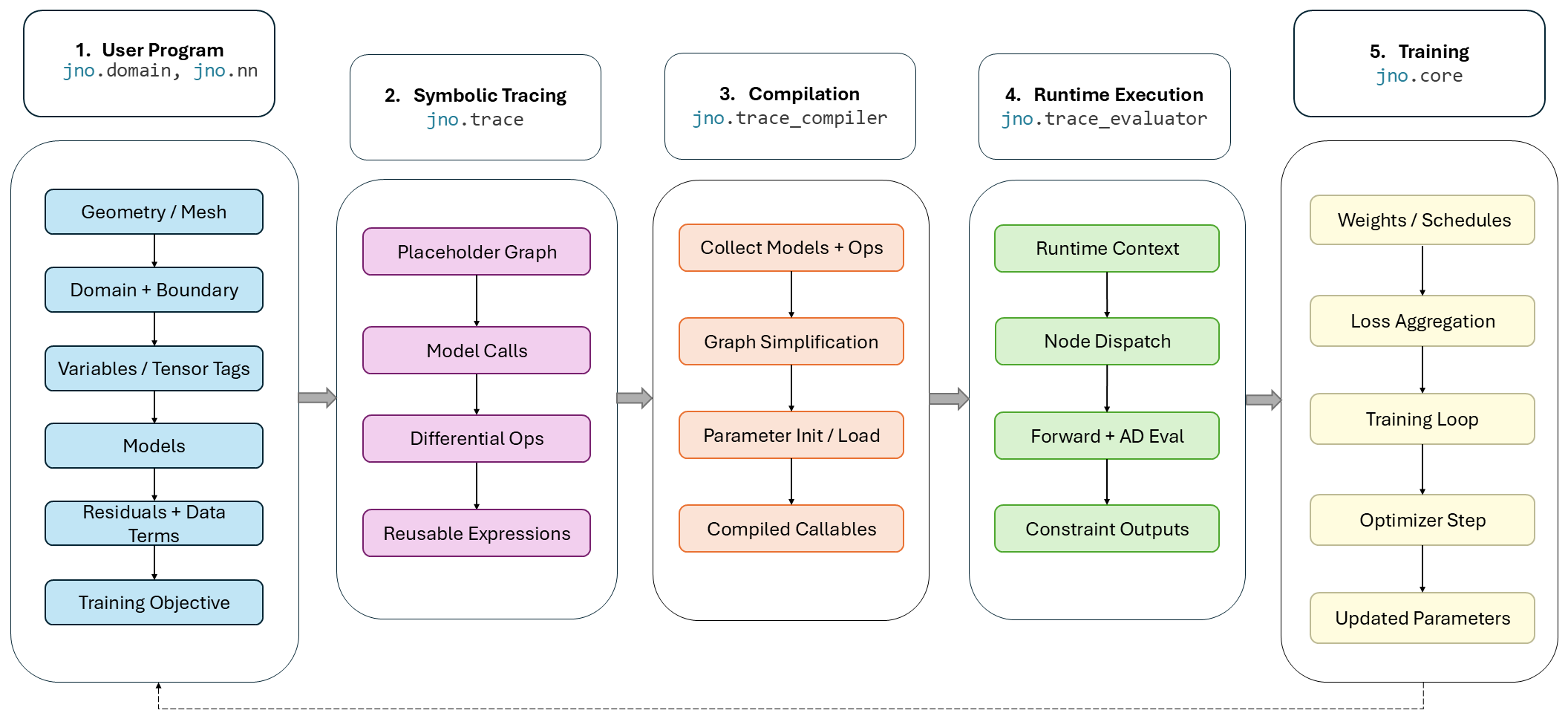}
  \caption{Main execution pipeline of jNO for model training.}
  \label{fig:jno-flowchart}
\end{figure}

In this section, we describe the central modules of jNO and their usage.
Figure~\ref{fig:jno-flowchart} summarizes the main traced execution pipeline for neural-operator, PINN and foundation-model training.

\subsection{Tracing system}

The tracing layer in jNO is implemented as a symbolic Domain-Specific Language (DSL) centered on \texttt{Placeholder} nodes. Arithmetic operators, comparison operators, slicing, reductions (for example \texttt{.mse}, \texttt{.mean}), and differential operators are overloaded to construct expression trees rather than execute eagerly. Many of these \texttt{Placeholder} operations are thin symbolic wrappers over \texttt{jax.numpy} primitives, so traced expressions stay close to standard JAX tensor semantics while deferring execution. In this model, objects such as \texttt{Variable}, \texttt{TensorTag}, \texttt{Literal}, \texttt{Constant}, \texttt{Model}, \texttt{Jacobian}, and \texttt{Hessian} all represent nodes in one deferred graph that is compiled and evaluated at a later stage.

Two design choices are important for robustness. First, placeholders use identity-based equality/hashing, so symbolic nodes can be reused safely across sets, dictionaries, and operation registries without accidental value-based collisions. Second, repeated callable expressions are normalized through \texttt{OperationDef}/\texttt{OperationCall}: users may write reusable equation fragments once and rebind them to different variable instances during evaluation.

Model invocations are first-class graph nodes through \texttt{Model} and \texttt{ModelCall}. Training controls (freezing, masking, optimizer attachment, LoRA activation~\citep{hu2022lora}, dtype selection, weight initialization, and per-group optimization settings) are attached to these model objects and therefore remain part of the same symbolic programming flow as PDE and data terms. In practice, this lets users express multi-model objectives directly in traced algebra, without introducing an additional orchestration layer.

Before execution, jNO performs common sub-expression elimination on the traced tree. Structurally identical subtrees (including function calls, operation calls, model calls, and derivative nodes) are canonicalized to shared nodes, reducing redundant work and improving compile-time and runtime efficiency. The same tree can also be inspected using \texttt{dump\_tree} for structural debugging.

Runtime evaluation is handled by \texttt{TraceEvaluator}, which dispatches node types through a handler table and evaluates a single batch context with explicit variable bindings. Differential operators support both automatic differentiation and finite-difference schemes, with finite-difference evaluation driven by mesh information carried in the domain object (cf.\ Section~\ref{ssc:domain}). The implementation treats temporal and spatial derivatives differently where needed, e.g., temporal derivatives via \texttt{\_\_time\_\_} context handling. In finite-difference mode, one can evaluate on mesh vertices and map values to general sampled points.

The tracing system also includes shape-level diagnostics: \texttt{trace\_\allowbreak shapes} records node-wise output shapes, and tracker nodes evaluate monitored expressions at intervals without adding to the loss. At the user level, \texttt{core.\allowbreak print\_shapes()} shows a readable path from model inputs to final loss terms, making dependency flow and shape propagation explicit for hybrid objectives.

\subsection{Domain and mesh workflow}\label{ssc:domain}

The \texttt{domain} class acts as a mesh-aware data organization layer between geometry definitions and traced expressions. It supports both mesh generation from constructors (via PyGmsh~\citep{schlomer2018pygmsh} or Gmsh~\citep{geuzaine2009gmsh}) and loading external meshes through \texttt{meshio} \citep{nschloe2024meshio}. Built-in constructors include common 1D, 2D, and 3D geometries (for example, line, rectangle, disk, L-shape, cube, structured rectangular grids, and multi-hole variants), and each geometry is exposed through physical tags such as \texttt{interior}, \texttt{boundary}, and side-specific subsets.

Internally, domain data are organized in two complementary stores: a mesh pool for full tagged point sets and a runtime context dictionary used by the solver. Sampling operations populate context arrays with explicit batch/time/point structure. Spatial tags are stored in a consistent shape family that supports both steady and time-dependent workflows, while temporal information is handled as a separate \texttt{\_\_time\_\_} context entry. This separation allows the tracing/evaluation code to treat space and time consistently across residual, data, and mixed operator objectives.

The API also supports batched neural operator training setups directly at the domain level. Repetition and merging operators (for example, scaling a domain by an integer or adding domains) are used to construct neural operator training data as multi-sample contexts with aligned tags. In parallel, tensor tags can be attached for parametric inputs and coefficients, enabling one program to carry both geometric coordinates and sample-wise conditioning tensors. This is central for neural operator training where each batch element may represent a different PDE parameterization or forcing field.

A minimal setup is shown in Listing~\ref{lst:domain-batch-rect}. Here, 500 rectangular domains are batched, interior and boundary coordinates are extracted explicitly, and one scalar parameter per rectangle is attached as a tensor tag. To remain compatible with jNO runtime conventions, attached tensor tags should follow the shape \texttt{(B, T, ...)}.

\begin{lstlisting}[caption={Batched rectangular domains with coordinate extraction and per-sample parameter tensor.},label={lst:domain-batch-rect}]
dom = 500 * jno.domain.rect(mesh_size=0.05, x_range=(0, 2), y_range=(0, 1))
x, y, _ = dom.variable("interior")
xb, yb, _ = dom.variable("boundary")

random_k = jax.random.uniform(jax.random.PRNGKey(0), shape=(500, 1, 1), minval=0.5, maxval=1.5)
k = dom.variable("k", random_k)
\end{lstlisting}

For derivative and residual workflows, \texttt{domain.py} precomputes mesh connectivity metadata (neighbors, element topology, nodal measures, and boundary indices) that are consumed by finite-difference pathways in the evaluator. Boundary normals are computed from mesh geometry (including handling of internal boundaries), and optional visibility/view-factor operators are provided for radiative boundary conditions, including multi-boundary enclosure assembly and cross-boundary factor blocks. These capabilities allow the same high-level DSL to address both standard PDE constraints and geometry-coupled boundary physics.

Finally, the domain layer includes explicit support for adaptive point management. Sampling hooks and per-tag resampling strategy registration are integrated into the runtime context model so that training can update point sets without changing symbolic equations. Together, these functionalities make jNO's \texttt{domain} class more than a geometry container: it is the bridge that aligns meshing, tagging, batching, parametric tensors, and physics-specific boundary structures with the traced optimization pipeline.

\subsection{Finite Element Method Workflow}

The finite element route in jNO extends the same traced programming model used for residual based and operator-learning workflows to weak form assembly. 
Rather than introducing a separate FEM specific front end, jNO keeps meshes, tags, variables, and symbolic expressions in one DSL and defers backend selection to assembly time. 
In this way, mesh-aware weak forms become another execution path of the same traced program, alongside pointwise residual evaluation, variational PINN training, operator learning, and space-time-dependent solver workflows.

The workflow begins from the domain layer described in Section~\ref{ssc:domain}. 
Domains are created from built-in geometries or loaded meshes and expose physical tags such as interior, boundary, and side-specific subsets. 
As in the rest of jNO, the domain object acts as the bridge between mesh data and traced expressions by storing tagged point sets and runtime context arrays in a unified structure. 
This design is important for the FEAX route because it allows weak form variables, quadrature data, and boundary metadata to remain aligned with the same tagging and batching conventions already used for residual and operator learning workflows.

After mesh construction, \texttt{init\_fem(...)} augments the domain with the FEAX~\citep{ichihara2025feax} data needed for weak-form assembly.
Internally, this step initializes the  FEAX mesh representation,  element metadata,  quadrature data, cached shape function tensors, and boundary condition metadata, while also registering variational sampling regions in the domain context.
In particular, the volume quadrature region is exposed through \texttt{fem\_gauss}, and tagged boundary quadrature regions are exposed through tags of the form \texttt{gauss\_<tag>}.
Because these regions are registered back into the domain, subsequent variable creation follows the same \texttt{domain.variable(...)} interface as elsewhere in jNO, with additional variational metadata attached where appropriate.

Weak forms are then written symbolically using the same traced algebra used throughout the library. 
The FEM route introduces generic trial and test symbols and keeps them inside the traced graph, rather than moving to a separate equation language. 
This preserves the central jNO design in which symbolic nodes, model calls, derivatives, and downstream execution remain part of one deferred program. 
As a result, weak form expressions can still be composed from traced derivatives, source terms, and boundary contributions in the same user-facing syntax, as illustrated in Listing~\ref{lst:weakform-jno-fem}.

At assembly time, the symbolic weak form is grouped into volume and tagged boundary contributions and lowered to a backend-neutral weak-form representation.
The same symbolic interface can be dispatched to \texttt{vpinn}, \texttt{fem\_system}, \texttt{fem\_residual}, or \texttt{fem\_time}.
For variational PINN workflows, the \texttt{vpinn} route replaces trial symbols by neural trial expressions and returns a differentiable grouped weak residual for training.
For steady linear finite element problems, the \texttt{fem\_system} route returns a JAX-native linear system \((A,b)\), while nonlinear steady problems can be routed through \texttt{fem\_residual}, which returns a residual operator with a Jacobian callable.
This allows assembled FEM operators to be passed to external linear or nonlinear solvers without changing the symbolic weak-form definition.

The transient FEM route extends the same mechanism to time-dependent weak forms.
When a weak form contains temporal derivatives of the trial field, \texttt{weak.assemble(...)} returns a solver-facing semidiscrete time block assembled through FEAX.
For first-order linear problems, the block represents a semi-discrete system of the form
\[
    M \dot{u} + A u = b(t),
\]
where the mass matrix, operator matrix, affine bias, and optional forcing callback are exposed in JAX-compatible form.
For nonlinear first-order problems, the block represents
\[
    M(t)\dot{u} + R(u,t) = 0,
\]
and provides callable mass, residual, and Jacobian functions.
The resulting block can be converted to a Diffrax-compatible explicit first-order system or to a FEAX time-integration pipeline, while still remaining accessible to external solver libraries.

This FEAX integration is important because it keeps the assembled finite element operators inside the JAX ecosystem.
Compared with a PETSc-centered assembly path, the FEAX route makes end-to-end JAX-differentiable workflows possible: weak-form assembly, residual evaluation, Jacobian construction, neural model coupling, and solver-in-the-loop computations can all remain compatible with JAX transformations.
At the same time, jNO does not prescribe a single solver backend.
The assembled systems and residual operators can be solved with JAX-native solvers such as Diffrax for time integration and Optimistix for nonlinear solves, or with external Python solvers such as SciPy when a conventional CPU-based reference solve is desired.
Thus, FEAX acts as the differentiable finite element lowering backend, while jNO keeps solver selection independent from the symbolic problem definition.

\begin{lstlisting}[caption={Tagged weak form assembly in the jNO FEAX-FEM route.},label={lst:weakform-jno-fem}]
dom = jno.domain.rect(mesh_size=0.05, x_range=(0, 1), y_range=(0, 1))
dom.init_fem(element_type="TRI3", quad_degree=2,
    bcs=[dom.dirichlet(["left","right"], 0.0), dom.neumann(["top", "bottom"])])

u, phi = dom.fem_symbols()

xg, yg, _ = dom.variable("fem_gauss")
xt, yt, _ = dom.variable("gauss_top")
xb, yb, _ = dom.variable("gauss_bottom")

du_dx = jnn.grad(u, xg);   du_dy = jnn.grad(u, yg); 
phi_x = jnn.grad(phi, xg); phi_y = jnn.grad(phi, yg)

weak = (du_dx * phi_x + du_dy * phi_y - source_f(xg, yg) * phi 
     - neumann_top(xt, yt) * phi - neumann_bottom(xb, yb) * phi)

A, b = weak.assemble( target="fem_system")
\end{lstlisting}

Listing~\ref{lst:weakform-jno-feax-time} shows the corresponding transient route.
The same weak-form syntax is used, but the temporal derivative causes the expression to be lowered into a semi-discrete FEAX time block rather than a steady linear system.

\begin{lstlisting}[caption={Transient weak form assembly into a FEM time block.},label={lst:weakform-jno-feax-time}]
dom = jno.domain.rect( mesh_size=0.05, x_range=(0, 1),  y_range=(0, 1), time=(0.0, 1.0, 1),)

dom.init_fem(
    element_type="TRI3",   quad_degree=2,
    bcs=[dom.dirichlet(["left", "right", "top", "bottom"], 0.0)],)

u, phi = dom.fem_symbols()

xg, yg, tg = dom.variable("fem_gauss", split=True)

u_t = jnn.grad(u, tg);  u_x = jnn.grad(u, xg);  u_y = jnn.grad(u, yg)
phi_x = jnn.grad(phi, xg);  phi_y = jnn.grad(phi, yg)

weak = u_t * phi + nu * (u_x * phi_x + u_y * phi_y)

block = weak.assemble( target="fem_time",  linear=True,  state0=u0, mode="implicit",)
diffrax_block = block.as_diffrax()
feax_pipeline = block.as_feax_pipeline(scheme="backward_euler")
\end{lstlisting}

Overall, the FEM route should therefore be understood as an integration layer between jNO's tracing system and FEAX-based differentiable finite element assembly.
The main contribution is not a new finite element formulation, but a unified workflow in which mesh-based weak forms, variational PINNs, operator models, steady FEM systems, nonlinear residual operators, and transient semidiscrete systems can be composed without changing the surrounding programming model.
This is consistent with the broader design goal of jNO: to avoid fragmented PDE learning software stacks by keeping domains, equations, models, assembly, and execution inside one symbolic interface.

\subsection{Model interface and foundation models}

jNO exposes all neural operator architectures through an external repository \cite{foundax}. Model categories include all common neural operator architectures~\citep{azizzadenesheli2024neural}: spectral methods (FNO, GeoFNO, PCNO), convolutional architectures (UNet, CNO, MgNO), attention-based methods (Transformer, PiT, CGPTNO, GNOT), branch-trunk decompositions (DeepONet), and point-based methods (PointNet). Critically, models imported from foundax \citep{foundax} are treated as ordinary operations within the traced program when wrapped via \texttt{jno.nn.\allowbreak wrap()}: they can be called, composed, and combined with arbitrary traced expressions just like any other function. Custom Equinox \citep{kidger2021equinox} modules can be treated the same way. This design makes it straightforward to use multiple models in a single program.

For foundation model workflows, jNO targets JAX-native translation of PDE foundation models that have predominantly appeared in PyTorch ecosystems. The project collaborates with translated model families including Poseidon, Walrus, PDEformer2, MPP, BCAT, DPOT and Morph, with model-specific implementations and pretrained checkpoints maintained in dedicated external repository \citep{foundax}. This decoupled structure allows fine-tuning and transfer workflows to use the same training and tuning interfaces as built-in operators, so users can compose foundation backbones with custom adaptation layers, physics constraints, or multi-model ensembles without switching frameworks.

Optimizers and training controls are attached to model objects before solving: each model in the traced constraints is assigned an optimizer via \texttt{model.optimizer()}, and controls include freeze/unfreeze, selective masking by pytree path, LoRA activation~\citep{hu2022lora}, and dtype selection (\texttt{model.dtype} supports \texttt{bfloat16}). A representative setup for initialization, optimizer assignment, masking, and LoRA activation is shown in Listing~\ref{lst:model-controls}. A more task-level operator example is shown in Listing~\ref{lst:model-deeponet}, where a DeepONet is configured to consume domain coordinates and a per-sample parameter tensor.

\begin{lstlisting}[caption={Example model-level control setup in jNO.},label={lst:model-controls}]
  model = jno.nn.wrap(foundax.poseidon(...))
  model.initialize("<path_to_model>.msgpack")
  model.optimizer(optax.adamw(optax.schedules.cosine(1e-3)))
  model.mask(boolean_param_mask).lora(rank=4, alpha=8)
\end{lstlisting}

\begin{lstlisting}[caption={DeepONet construction },label={lst:model-deeponet}]
net = jno.nn.wrap(foundax.deeponet(
  n_sensors=1,
  coord_dim=2,
  basis_functions=32,
  hidden_dim=128,
))
net.optimizer(
  optax.adam(
    learning_rate=optax.schedules.cosine_decay_schedule(
      init_value=1e-3, decay_steps=10_000, alpha=1e-5
    )
  )
)
\end{lstlisting}

LoRA is supported for both Equinox \citep{kidger2021equinox} and legacy Flax \citep{heek2024flax} models, with training restricted to the adapted parameters when enabled.

\subsection{Forward pass and boundary-condition encoding}

Once a model has been instantiated, jNO treats the forward pass itself as part of the traced symbolic program. This is also the natural place to encode hard boundary conditions through output transformations before residual terms are formed. Listing~\ref{lst:forward-bc} shows a typical pattern in which the raw model output is multiplied by a boundary-vanishing envelope and then inserted into a PDE residual.

\begin{lstlisting}[caption={Exemplary forward pass with hard boundary condition enforcement (zero Dirichlet boundary conditions on {$[0,1]^2$}) and PDE residual (for PDE $k\,\Delta u+1=0$).},label={lst:forward-bc}]
u = net(k, jno.np.concat([x, y], axis=-1)) * x * (1 - x) * y * (1 - y)
pde = k * (u.dd(x) + u.dd(y)) + 1.0
\end{lstlisting}

\subsection{Runtime and execution features in \texttt{core.py}}

When constructing \texttt{crux = jno.core(...)} and then calling \texttt{crux.solve(...)}, the runtime in \texttt{jno/core.py} provides multi-device execution, memory and performance controls, and low-overhead training loops. \texttt{core} configures a 2-axis device mesh \texttt{(batch, model)} for data and model parallelism with a data-parallel fallback when the requested mesh is unavailable. Constraints are compiled into one combined function so XLA can eliminate shared subexpressions, and the step function is JIT compiled with explicit shardings and buffer donation. The training path enables persistent XLA cache storage and a short warmup phase. \texttt{solve()} exposes memory controls (\texttt{checkpoint\_gradients}, \texttt{offload\_data}), throughput fusion via \texttt{inner\_steps}, resampling at outer-step boundaries, and bounded profiling via \texttt{profile}. Listing~\ref{lst:runtime-solve} shows the minimal pattern for assembling the solver, inspecting traced shapes, and launching optimization.

\begin{lstlisting}[caption={Core solver construction and training launch.},label={lst:runtime-solve}]
dir = jno.setup("./runs/test")
crux = jno.core(constraints=[pde.mse], domain=dom, mesh=(1, 1)).print_shapes()
crux.solve(epochs=1_000, batchsize=32).plot(f"{dir}/training.png")
\end{lstlisting}

\subsection{Training and tuning}

The \texttt{core} solver compiles traced expressions into one JIT-compiled training program, so XLA-level common sub-expression elimination applies uniformly across objectives. The \texttt{solve()} interface mirrors the implementation in \texttt{jno.core}: it accepts \texttt{epochs}, \texttt{batchsize} (\texttt{None} for full-batch), \texttt{checkpoint\_gradients} for rematerialization, \texttt{offload\_data} to stream batches from host memory, \texttt{inner\_steps} for fused steps, \texttt{min\_consecutive} for constraint evaluation windows, and optional \texttt{profile} tracing. 

Hyperparameter tuning is supported through \texttt{ArchSpace}, which defines search ranges for categorical choices (via \texttt{unique}), continuous floats (via \texttt{float\_range} with optional log-scale sampling), and discrete integers (via \texttt{int\_range}). Search spaces are categorized into architecture parameters (layer counts, activation functions, mode counts) and training parameters (optimizer choice, learning rate, constraint weights). Grid search is supported natively through \texttt{ArchSpace.grid()}, while gradient-free optimization via Nevergrad \citep{facebook2018nevergrad} allows efficient search in high-dimensional or mixed-type spaces. The tuning interface is integrated with the core training loop, so the same traced program can be evaluated across many hyperparameter candidates without redefining constraints or reloading domain data.

\subsection{Experiment persistence and shareable artifacts}

jNO provides object-level persistence via the public API \texttt{jno.save} and \texttt{jno.load}. Three object types are supported: \texttt{core} solver states (including trained parameters, optimizer state, and training history), \texttt{domain} objects (with mesh and context data), and exported \texttt{IREEModel} wrappers for ahead-of-time compiled inference \citep{openxla2019iree}. Serialization uses \texttt{cloudpickle} \citep{cloudpipe2024cloudpickle}, so Python closures and module references are preserved.

For reproducibility and exchange, the API supports optional RSA-signed artifacts via \texttt{pylotte} \citep{alpamayo2026pylotte}. When keys are provided, \texttt{jno.save} writes a payload plus detached \texttt{.sig}, and \texttt{jno.load} verifies integrity; key paths can be set in \texttt{.jno.toml} (project) or \texttt{\textasciitilde/.jno/config.toml} (user). Unsigned and pylotte-signed files remain readable for backward compatibility.

In practice, this allows a trained experiment state to be moved as a single serialized file for local reuse, and optionally accompanied by a signature for verifiable sharing workflows. 




\section{Research impact statement}

jNO is intended as a reusable JAX-native research software stack for neural operator and PDE-constrained learning workflows that would otherwise be split across separate frameworks. Its impact is reflected in three concrete ways. First, the library is packaged for installation through PyPI, which lowers the barrier for reuse in reproducible computational workflows. Second, the repository includes automated tests and executable examples that serve as reference implementations for operator-learning and physics-informed training scenarios. Third, the project acts as the unified JAX-native backend for related PDE foundation-model families \citep{foundax}, thereby enhancing the further development of these models~\citep{medvedev2026physicsinformed}.

We emphasise that jNO is the first consolidated framework enabling systematic comparison and compositional workflows across multiple foundation models within a single execution model. Rather than requiring researchers to maintain separate codebases for each model family, jNO standardizes training controls, parameter-level optimization, and physics-informed fine-tuning through one traced interface. This lowers entry barriers for foundation-model research and enables hybrid workflows such as ensemble methods, progressive transfer learning, and multi-model constraints that were previously difficult to implement consistently.

Furthermore, the JAX-native design provides inherent performance advantages for JIT-compiled training and inference. The traced program compiles to a single XLA graph, enabling aggressive sub-expression elimination, memory optimization, and multi-device parallelism without user-level threading or device management. For researchers accustomed to PyTorch's eager execution, this trade-off (forward specification for structured compilation) yields measurable speedups in larger-scale operator and foundation-model training.

Together, these distribution, validation, and extension pathways provide credible research significance by making advanced neural-operator and foundation-model workflows more accessible and efficient within the JAX ecosystem.

\section{Quality control}

jNO uses automated checks and executable examples to support reliability across both operator-learning and hybrid workflows. The test suite includes unit tests for the tracing DSL (\texttt{test\_trace.py}, \texttt{test\_trace\_evaluator.py}), domain and geometry operations (\texttt{test\_domain\_geometries.py}, \texttt{test\_domain\_load\_mesh.py}), derivative operators, model architectures, adaptive resampling, and cross-framework integration. Additional coverage includes configuration management, signed save/load workflows, multi-device training, and IREE ahead-of-time compilation \citep{openxla2019iree}. Integration tests validate end-to-end training workflows with traced constraints, multi-model objectives, and solver state serialization.

The repository uses pytest \citep{pytestdev2009pytest} with custom markers for test categorization: \texttt{slow} for computationally intensive tests, \texttt{integration} for multi-component workflows, \texttt{gpu} for accelerator-required tests, and \texttt{serial} for tests that must run sequentially. Fixtures in \texttt{conftest.py} provide deterministic RNG keys and mock domain objects for consistent test behavior. Quick validation is also possible by running example scripts and confirming expected training-history outputs and plots.

\section{Availability}

\subsection{Operating system}

Linux, macOS, Windows (via WSL2 for CUDA support). Minimum: Ubuntu 20.04, macOS 11+, Windows 10+.

\subsection{Programming language}

Python (requires \texttt{>=3.11,<3.14} in the published package metadata).

\subsection{Dependencies}

Core dependencies in \texttt{pyproject.toml} are \texttt{jax} \citep{bradbury2018jax}, \texttt{equinox} \citep{kidger2021equinox}, \texttt{maskx} \citep{maskx2026maskx}, \texttt{optax} \citep{deepmind2020optax}, \texttt{pygmsh} \citep{schlomer2018pygmsh}, \texttt{cloudpickle} \citep{cloudpipe2024cloudpickle}, and \texttt{einops} \citep{arogozhnikov2022einops}. Optional extras provide CUDA-enabled JAX, development and testing tools, IREE integration \citep{openxla2019iree}, and foundation model adapters.

\subsection{List of contributors}





\begin{itemize}
  \item Leon Armbruster (Fraunhofer IISB, Germany): Software development, documentation, testing, writing.
  \item Rathan Ramesh (Fraunhofer IISB, Germany): FEM Integration.
  \item Georg Kruse (Fraunhofer IISB, Germany): Testing, writing, supervision.
  \item Christopher Straub (Fraunhofer IISB, Germany): Supervision, writing.
\end{itemize}

\subsection{Software location}

\textbf{Code repository}

\begin{description}[noitemsep,topsep=0pt]
  \item[Name:] GitHub (FhG-IISB/jNO)
  \item[Identifier:] \url{https://github.com/FhG-IISB/jNO}
  \item[PyPI:] \url{https://pypi.org/project/jax-neural-operators/}
  \item[Licence:] EPL-2.0 License
  \item[Date published:] March 26, 2026
\end{description}

\vspace{3mm}

\begin{description}[noitemsep,topsep=0pt]
  \item[Name:] GitHub (FhG-IISB/foundax)
  \item[Identifier:] \url{https://github.com/FhG-IISB/foundax}
  \item[PyPI:] \url{https://pypi.org/project/foundax/}
  \item[Licence:] EPL-2.0 License
  \item[Date published:] April 16, 2026
\end{description}

\subsection{Language}

English (documentation, comments, API) and Python (implementation).

\section{Reuse potential}

jNO is reusable wherever researchers need to combine operator-learning models and PDE structure without maintaining separate code stacks. The same traced programming model supports paired-data operator learning, physics-informed regularization, mesh-based residual evaluation, and foundation model fine-tuning. The project is also structured for extension: contributors can add models, operators, domain constructors, and resampling or tuning strategies without changing the core user-facing language.

\begin{itemize}
	\item \textbf{Scientific machine learning}: Users can train neural operators on paired datasets and add physics constraints for regularization or adaptation using the same traced syntax.
  \item \textbf{General domains}: The same programming model applies to fluid, diffusion, wave, and coupled systems when represented as operator maps and/or residual terms.
  \item \textbf{Foundation model transfer}: Translated JAX foundation models (Poseidon \cite{herde2024poseidon}, Walrus \cite{mccabe2025walrus}, PDEformer2 \cite{pdeformer2}, MPP \cite{mccabe2024multiple}, Morph \cite{rautela2025morph}), BCAT \cite{liu2025bcat} and DPOT \cite{hao2024dpot} can be incorporated into fine-tuning workflows while preserving the same solver and tuning interfaces.
  \item \textbf{Extension}: Contributors can add models, operators, domain constructors, and resampling/tuning strategies without changing the core user-facing language.
\end{itemize}

For contributions and support, please use the project issue tracker and discussions in the main repository \url{https://github.com/FhG-IISB/jNO}.

\section*{AI usage disclosure}

GitHub Copilot was used to assist with drafting and revising parts of the manuscript text and repository automation related to the paper submission workflow. The author reviewed, edited, and validated all AI-assisted output and remains fully responsible for the technical claims, wording, citations, and final submitted materials.

\section*{Acknowledgements}


We would also like to thank the members of the AI-Augmented Research Group Janhavi Halgarkar, Vlad Medvedev, Philipp Brendel, and Rodrigo Coehlo, for their input.

\section*{Competing interests}

The author declares that there are no competing interests.

\bibliographystyle{plainnat}
\bibliography{references}

@article{raissi2019physics,
  title={Physics-informed neural networks: A deep learning framework for solving forward and inverse problems involving nonlinear partial differential equations},
  author={Raissi, Maziar and Perdikaris, Paris and Karniadakis, George E},
  journal={Journal of Computational Physics},
  volume={378},
  pages={686--707},
  year={2019},
  publisher={Elsevier},
  doi={10.1016/j.jcp.2018.10.045},
  url={https://doi.org/10.1016/j.jcp.2018.10.045}
}

@misc{ichihara2025feax,
  author       = {Ichihara, Naruki},
  title        = {{FEAX}: Finite Element Analysis with {JAX}},
  year         = {2025},
  howpublished = {\url{https://github.com/Naruki-Ichihara/feax}},
  note         = {Software library. Accessed: 2026-04-29}
}

@inproceedings{li2021fno,
  title={Fourier neural operator for parametric partial differential equations},
  author={Li, Zongyi and Kovachki, Nikola and Azizzadenesheli, Kamyar and Liu, Burigede and Bhattacharya, Kaushik and Stuart, Andrew and Anandkumar, Anima},
    booktitle={Proc.\ 9th International Conference on Learning Representations (ICLR, 2021).},
    year={2021},
    url={https://arxiv.org/abs/2010.08895}
}

@article{azizzadenesheli2024neural,
  title={Neural Operators for Accelerating Scientific Simulations and Design},
  author={Kamyar Azizzadenesheli and Nikola Kovachki and Zongyi Li and Miguel Liu-Schiaffini and Jean Kossaifi and Anima Anandkumar},
  journal={Nat.\ Rev.\ Phys.},
  volume={6},
  pages={320--328},
  year={2024},
  doi={10.1038/s42254-024-00712-5},
  url={https://doi.org/10.1038/s42254-024-00712-5}
  }

@article{lu2021deeponet,
  title={Learning nonlinear operators via DeepONet based on the universal approximation theorem of operators},
  author={Lu, Lu and Jin, Pengzhan and Pang, Guofei and Zhang, Zhongqiang and Karniadakis, George Em},
  journal={Nature Machine Intelligence},
  volume={3},
  number={3},
  pages={218--229},
  year={2021},
  publisher={Nature Publishing Group},
  doi={10.1038/s42256-021-00302-5},
  url={https://doi.org/10.1038/s42256-021-00302-5}
}

@inproceedings{
hu2022lora,
title={Lo{RA}: Low-Rank Adaptation of Large Language Models},
author={Edward J Hu and yelong shen and Phillip Wallis and Zeyuan Allen-Zhu and Yuanzhi Li and Shean Wang and Lu Wang and Weizhu Chen},
booktitle={International Conference on Learning Representations},
year={2022},
url={https://openreview.net/forum?id=nZeVKeeFYf9}
}

@inproceedings{
medvedev2026physicsinformed,
title={Physics-informed fine-tuning of foundation models for partial differential equations},
author={Vlad Medvedev and Leon Armbruster and Christopher Straub and Georg Kruse and Andreas Rosskopf},
booktitle={AI{\&}PDE: ICLR 2026 Workshop on AI and Partial Differential Equations},
year={2026},
url={https://openreview.net/forum?id=Ytpw2sXLmw}
}

@software{bradbury2018jax,
  author = {James Bradbury and Roy Frostig and Peter Hawkins and Matthew James Johnson and Chris Leary and Dougal Maclaurin and George Necula and Adam Paszke and Jake Vander{P}las and Skye Wanderman-{M}ilne and Qiao Zhang},
  title = {{JAX}: composable transformations of {P}ython+{N}um{P}y programs},
  url = {http://github.com/jax-ml/jax},
  version = {0.3.13},
  year = {2018},
}

@article{lu2021deepxde,
  title={DeepXDE: A deep learning library for solving differential equations},
  author={Lu, Lu and Meng, Xuhui and Mao, Zhiping and Karniadakis, George Em},
  journal={SIAM Review},
  volume={63},
  number={1},
  pages={208--228},
  year={2021},
  publisher={SIAM},
  doi={10.1137/19M1274067},
  url={https://doi.org/10.1137/19M1274067}
}

@software{hennigh2021nvidia,
  title={NVIDIA PhysicsNeMo: An open-source framework for physics-based deep learning in science and engineering},
  author={{PhysicsNeMo Contributors}},
  year={2023},
  month={2},
  day={24},
  url={https://github.com/NVIDIA/physicsnemo},
  note={If you use this software, please cite it as below. CFF version 1.2.0}
}

@article{zubov2021neuralpde,
  title={NeuralPDE: Automatic Physics-Informed Neural Networks (PINN) with Julia},
  author={Zubov, Kirill and McCarthy, Zoe and Ma, Yingbo and Calisto, Francesco and Pagliarino, Valerio and Azeglio, Simone and Bottero, Luca and Lujan, Emmanuel and Sulzer, Valentin and Bharambe, Ashutosh and others},
  journal={arXiv preprint arXiv:2107.09443},
  year={2021},
  url={https://arxiv.org/abs/2107.09443}
}

@article{geuzaine2009gmsh,
  title={Gmsh: A 3-D finite element mesh generator with built-in pre-and post-processing facilities},
  author={Geuzaine, Christophe and Remacle, Jean-Fran{\c{c}}ois},
  journal={International Journal for Numerical Methods in Engineering},
  volume={79},
  number={11},
  pages={1309--1331},
  year={2009},
  publisher={Wiley Online Library},
  doi={10.1002/nme.2579},
  url={https://doi.org/10.1002/nme.2579}
}

@misc{kharazmi2019variationalphysicsinformedneuralnetworks,
      title={Variational Physics-Informed Neural Networks For Solving Partial Differential Equations}, 
      author={E. Kharazmi and Z. Zhang and G. E. Karniadakis},
      year={2019},
      eprint={1912.00873},
      archivePrefix={arXiv},
      primaryClass={cs.NE},
      url={https://arxiv.org/abs/1912.00873}, 
}

@article{kidger2021equinox,
    author={Patrick Kidger and Cristian Garcia},
    title={{E}quinox: neural networks in {JAX} via callable {P}y{T}rees and filtered transformations},
    year={2021},
    journal={Differentiable Programming workshop at Neural Information Processing Systems 2021}
    ,url={https://arxiv.org/abs/2111.00254}
}

@software{deepmind2020optax,
  title={Optax: A gradient processing and optimization library for JAX},
  author={Babuschkin, Igor and Baumli, Kate and Bell, Alison and Bhupatiraju, Surya and Bruce, Jake and Buchlovsky, Peter and Budden, David and Cai, Trevor and Clark, Aidan and Danihelka, Ivo and others},
  year={2020},
  url={https://github.com/deepmind/optax}
}

@software{heek2024flax,
  author = {Jonathan Heek and Anselm Levskaya and Avital Oliver and Marvin Ritter and Bertrand Rondepierre and Andreas Steiner and Marc van {Z}ee},
  title = {{F}lax: A neural network library and ecosystem for {JAX}},
  url = {http://github.com/google/flax},
  version = {0.12.5},
  year = {2024},
}

@software{schlomer2018pygmsh,
  title={pygmsh: A Python frontend for Gmsh},
  author={Schl{\"o}mer, Nico and others},
  year={2018},
  doi={10.5281/zenodo.1173105},
  url={https://github.com/meshpro/pygmsh}
}

@software{alpamayo2026pylotte,
  title={pylotte: RSA signing and verification utilities for Python artifacts},
  author={{Alpamayo Solutions}},
  year={2026},
  url={https://github.com/alpamayo-solutions/pylotte}
}

@software{predictive2024jaxpi,
  title={JAX-PI: Physics-informed neural networks in JAX},
  author={{Predictive Intelligence Lab}},
  year={2024},
  howpublished={\url{https://github.com/PredictiveIntelligenceLab/jaxpi}},
}

@software{nschloe2024meshio,
  title={meshio: Tools for reading and writing mesh data},
  author={Schl{\"o}mer, Nico and contributors},
  year={2024},
  howpublished={\url{https://github.com/nschloe/meshio}},
}

@misc{facebook2018nevergrad,
    author = {J. Rapin and O. Teytaud},
    title = {{Nevergrad - A gradient-free optimization platform}},
    year = {2018},
    publisher = {GitHub},
    journal = {GitHub repository},
    howpublished = {\url{https://GitHub.com/FacebookResearch/Nevergrad}},
}

@software{cloudpipe2024cloudpickle,
  title={cloudpickle: Extended pickling support for Python objects},
  author={{cloudpipe and contributors}},
  year={2024},
  howpublished={\url{https://github.com/cloudpipe/cloudpickle}},
}

@inproceedings{arogozhnikov2022einops,
    title={Einops: Clear and Reliable Tensor Manipulations with Einstein-like Notation},
    author={Alex Rogozhnikov},
    booktitle={International Conference on Learning Representations},
    year={2022},
    url={https://openreview.net/forum?id=oapKSVM2bcj}
}

@misc{maskx2026maskx,
  title={maskx: Mask algebra for selecting and combining JAX PyTree leaves},
  author={Armbruster, Leon},
  year={2026},
  howpublished={\url{https://pypi.org/project/maskx/}},
  note={Python package}
}

@misc{pytestdev2009pytest,
  title={pytest: A framework for testing Python code},
  author={{pytest-dev and contributors}},
  year={2009},
  howpublished={\url{https://github.com/pytest-dev/pytest}},
  note={Software repository}
}

@misc{openxla2019iree,
  title={IREE: End-to-end MLIR compiler and runtime for machine learning},
  author={{iree-org and contributors}},
  year={2019},
  howpublished={\url{https://github.com/iree-org/iree}},
  note={Software repository}
}

@misc{foundax,
  title={foundax},
  author={Leon Armbruster and contributors},
  year={2026},
  howpublished={\url{https://github.com/FhG-IISB/foundax}},
  note={Software repository}
}

@misc{mccabe2025walrus,
      title={Walrus: A Cross-Domain Foundation Model for Continuum Dynamics}, 
      author={Michael McCabe and Payel Mukhopadhyay and Tanya Marwah and Bruno Regaldo-Saint Blancard and Francois Rozet and Cristiana Diaconu and Lucas Meyer and Kaze W. K. Wong and Hadi Sotoudeh and Alberto Bietti and Irina Espejo and Rio Fear and Siavash Golkar and Tom Hehir and Keiya Hirashima and Geraud Krawezik and Francois Lanusse and Rudy Morel and Ruben Ohana and Liam Parker and Mariel Pettee and Jeff Shen and Kyunghyun Cho and Miles Cranmer and Shirley Ho},
      year={2025},
      eprint={2511.15684},
      archivePrefix={arXiv},
      primaryClass={cs.LG},
      url={https://arxiv.org/abs/2511.15684}, 
}

@misc{herde2024poseidon,
      title={Poseidon: Efficient Foundation Models for PDEs}, 
      author={Maximilian Herde and Bogdan Raonić and Tobias Rohner and Roger Käppeli and Roberto Molinaro and Emmanuel de Bézenac and Siddhartha Mishra},
      year={2024},
      eprint={2405.19101},
      archivePrefix={arXiv},
      primaryClass={cs.LG}
}

@article{rautela2025morph,
  title={MORPH: PDE Foundation Models with Arbitrary Data Modality},
  author={Rautela, Mahindra Singh and Most, Alexander and Mansingh, Siddharth and Love, Bradley C and Biswas, Ayan and Oyen, Diane and Lawrence, Earl},
  journal={arXiv preprint arXiv:2509.21670},
  year={2025}
}

@inproceedings{
  mccabe2024multiple,
  title={Multiple Physics Pretraining for Spatiotemporal Surrogate Models},
  author={Michael McCabe and Bruno R{\'e}galdo-Saint Blancard and Liam Holden Parker and Ruben Ohana and Miles Cranmer and Alberto Bietti and Michael Eickenberg and Siavash Golkar and Geraud Krawezik and Francois Lanusse and Mariel Pettee and Tiberiu Tesileanu and Kyunghyun Cho and Shirley Ho},
  booktitle={The Thirty-eighth Annual Conference on Neural Information Processing Systems},
  year={2024},
  url={https://openreview.net/forum?id=DKSI3bULiZ}
}

@misc{pdeformer2,
      title={PDEformer-2: A Versatile Foundation Model for Two-Dimensional Partial Differential Equations},
      author={Zhanhong Ye and Zining Liu and Bingyang Wu and Hongjie Jiang and Leheng Chen and Minyan Zhang and Xiang Huang and Qinghe Meng. Jingyuan Zou and Hongsheng Liu and Bin Dong},
      year={2025},
      eprint={2507.15409},
      archivePrefix={arXiv},
      primaryClass={math.NA},
      url={https://arxiv.org/abs/2507.15409},
}

@article{liu2025bcat,
  title={{BCAT}: A Block Causal Transformer for PDE Foundation Models for Fluid Dynamics},
  author={Yuxuan Liu and Jingmin Sun and Hayden Schaeffer},
  journal={arXiv preprint arXiv:2501.18972},
  year={2025},
  url={https://github.com/bootjp/bcat},
}

@article{hao2024dpot,
  title={DPOT: Auto-Regressive Denoising Operator Transformer for Large-Scale PDE Pre-Training},
  author={Hao, Zhongkai and Su, Chang and Liu, Songming and Berner, Julius and Ying, Chengyang and Su, Hang and Anandkumar, Anima and Song, Jian and Zhu, Jun},
  journal={arXiv preprint arXiv:2403.03542},
  year={2024},
  url={https://github.com/HaoZhongkai/DPOT}
}

\end{document}